\documentclass[runningheads]{llncs}
\usepackage{graphicx}
\usepackage{color}
\usepackage{amsmath}
\usepackage{amsfonts}       
\usepackage{algorithm}
\usepackage{algorithmic}
\usepackage[export]{adjustbox}
\usepackage{booktabs}       
\usepackage{multirow}
\usepackage{wrapfig}
\usepackage{hyperref}

\DeclareMathOperator{\vol}{Vol}
\DeclareMathOperator{\area}{Area}

\DeclareMathOperator*{\argmin}{arg\,min}

\newcommand{\R}{\mathbb{R}}
\newcommand{\IR}{\mathbb{I}\mathbb{R}}
\newcommand{\E}{\mathbb{E}}
\newcommand{\rd}{\delta}

\newcommand{\RE}{\mathbb{R}}
\newcommand{\calX}{\mathcal{X}}
\newcommand{\calB}{\mathcal{B}}
\newcommand{\calF}{\mathcal{F}}

\newcommand{\calP}{\bar{p}}
\newcommand{\calV}{\mathcal{V}}
\newcommand{\calU}{\mathcal{U}}
\newcommand{\calR}{\mathcal{R}}
\newcommand{\calS}{\mathcal{S}}

\newcommand{\chij}[1]{\chi(t_j,#1)}
\newcommand{\chit}[1]{\chi(t,#1)}

\newcommand{\hatM}{\hat{M}}

\newcommand{\Lp}{L}

\begin{document}
\title{Robustness Analysis of Continuous-Depth Models with Lagrangian Techniques}
\titlerunning{Robustness Analysis of Continuous-Depth Models}
%
\author{Sophie A. Neubauer (née Gruenbacher)\inst{1}\thanks{Correspondence to: \texttt{sophie.neubauer@tuwien.ac.at}~~ \newline Code: \url{https://github.com/DatenVorsprung/GoTube}
    } \and
Radu Grosu\inst{1}}
\authorrunning{S. A. Neubauer et al.}
%
\institute{Technische Universit\"at Wien (TU Wien), Vienna, Austria}
\maketitle              
\begin{abstract}
This paper presents, in a unified fashion, deterministic as well as statistical Lagrangian-verification techniques. They formally quantify the behavioral robustness of any time-continuous process, formulated as a continuous-depth model. To this end, we review LRT-NG, SLR, and GoTube, algorithms for constructing a tight reachtube, that is, an over-approximation of the set of states reachable within a given time-horizon, and provide guarantees for the reachtube bounds. We compare the usage of the variational equations, associated to the system equations, the mean value theorem, and the Lipschitz constants, in achieving deterministic and statistical guarantees. In LRT-NG, the Lipschitz constant is used as a bloating factor of the initial perturbation, to compute the radius of an ellipsoid in an optimal metric, which over-approximates the set of reachable states. In SLR and GoTube, we get statistical guarantees, by using the Lipschitz constants to compute local balls around samples. These are needed to calculate the probability of having found an upper bound, of the true maximum perturbation at every timestep. Our experiments demonstrate the superior performance of Lagrangian techniques, when compared to LRT, Flow*, and CAPD, and illustrate their use in the robustness analysis of various continuous-depth models.

\end{abstract}
%
%
\section{Introduction}
Due to the revival of neural ordinary differential equations (Neural ODEs)~\cite{neuralODEs}, modern cyber-physical systems (CPS) increasingly use deep-learning systems powered by continuous-depth models, where the dynamics of the hidden states are defined by an ordinary differential equation (ODE) and the output is a function of the solution of the ODE at a given time. They are used within the cyber part of the CPS responsible for state-estimation, planning, and (adaptive) optimal control, of the physical part of the CPS. As the use of continuous-depth models on real-world applications increases \cite{finlay2020train,lechner2020neural,erichson2020lipschitz,lechner2020learning,hasani2020natural}, so does the importance of ensuring their safety through the use of verification techniques.

Since all these networks represent nonlinear systems of ordinary differential equations, it is impossible, that is, undecidable, to exactly predict their behavior, as they do not have a closed-form solution. This is very problematic because safety is an important concern in many of such systems, as for example, smart mobility, industry 4.0, or smart health-care. Fortunately, it is possible to approximate this behavior. Robustness analysis of continuous-depth models, can be seen as a special case of reachability analysis of nonlinear ordinary differential equations (ODEs), as it measures the ability to resist change in the input values. 

In this case, the problem is how to over-approximate the system dynamics, and thus the behaviour of the system, in as tight a way as possible, so that one can rely upon and use the huge potential of these continuous-depth models even in safety-critical systems, when it comes to difficult tasks. To avoid false positives when looking for intersections of the systems with unsafe regions, it is crucial to have as-tight-as-possible reachtubes. Otherwise it would e.g. predict that a car driven by a controller would cause a crash even if the neural network controller is behaving perfectly and never causes a crash. Such wide reachtubes are thus not useful for actually putting continuous depth-models into operation.
%
\begin{figure}
\centering
\includegraphics[width=0.5\textwidth]{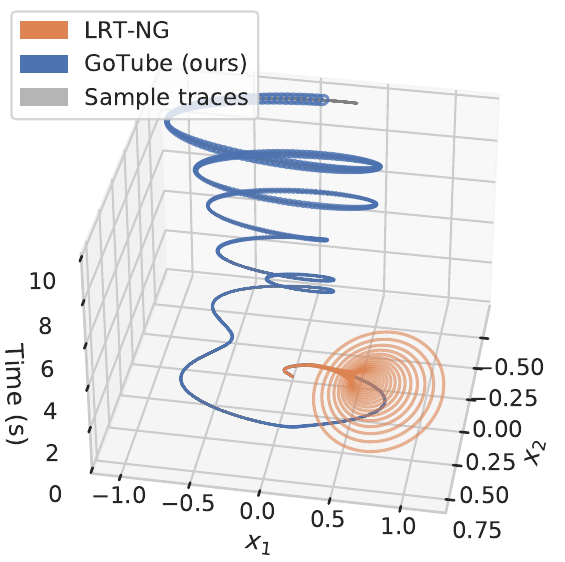}
\caption{Reachtubes of LRT-NG~\cite{gruenbacher2020lagrangian} and GoTube~\cite{Gruenbacher2021GoTube} for a CT-RNN controlling Cart-Pole-v1 environment. LRT~\cite{Cyranka2017}, CAPD~\cite{capd}, and Flow*~\cite{flowstar} failed on this benchmark.}
\label{fig:intro}
\end{figure}
%

In this work, we focus on the importance of the variational equation used in our tools. We review our conservative, set-based reachability tool LRT-NG~\cite{gruenbacher2020lagrangian}, our non-conservative, stochastic theory SLR~\cite{Gruenbacher2021verification}, and scalable statistical tool GoTube~\cite{Gruenbacher2021GoTube}. The stochasticity of SLR is only introduced through the algorithm, we are not looking at stochastic dynamical systems.
Deterministic verification approaches ensure conservative bounds~\cite{flowstar,gowal2018effectiveness,mirman2018differentiable,bunel2020lagrangian,capd}, but often sacrifice speed and accuracy \cite{ehlers2017formal}, especially due to the wrapping effect caused by interval arithmetic, and thus scalability; see CAPD, Flow*, LRT, and LRT-NG in Figs.~\ref{fig:intro},\,\ref{fig:three graphs}. Statistical methods, on the other hand, only ensure a weaker notion of conservativeness in the form of confidence intervals (statistical bounds). This, however, allows them to achieve much more accurate and faster verification algorithms that scale up to much larger dynamical systems~\cite{probreach,gp}.

We compared LRT-NG and GoTube with LRT, Flow*, and CAPD, on a comprehensive set of benchmarks, including continuous-depth models. Our results  show that LRT-NG is very competitive with both Flow* and CAPD. Moreover, it is the only conservative tool able to handle the continuous-depth models. GoTube substantially outperforms all state-of-the-art verification tools, in terms of the size of the initial ball, time-horizon, task completion, and scalability.
\begin{table*}[t]
\small
\centering
\caption{Related work on the reachability analysis of continuous-time systems. Determ. is the abbreviation for deterministic, and no indicates a statistical method. The table content presented here is partially reproduced from \cite{Gruenbacher2021verification}.}
\begin{adjustbox}{width=\columnwidth}
\begin{tabular}{l|c|c|c|c}
\toprule
\textbf{Technique} & \textbf{Determ.} & \textbf{Parallel} & \textbf{wrapping} & \textbf{Arbitrary }\\
 & &  & \textbf{effect} & \textbf{Time-horizon}\\
\midrule
LRT \cite{Cyranka2017} (Ours) with Infinitesimal strain theory & yes & no & yes & no \\
CAPD \cite{capd} implements Lohner algorithm & yes & no &  yes & no  \\
Flow-star \cite{flowstar} with Taylor models & yes & no  & yes & no  \\
$\delta$-reachability \cite{deltadecidable} with approximate satisfiability & yes & no & yes & no \\
C2E2 \cite{c2e2} with discrepancy functions & yes & no  & yes & no \\
LDFM \cite{fansimul} by simulation, matrix measures & yes & yes  & no & no \\
TIRA \cite{tira} with second-order sensitivity & yes & yes & no & no \\
Isabelle/HOL \cite{isabelle} with proof-assistant & yes & no  & yes & no \\
Breach \cite{breach,donze} by simulation & yes & yes  & no & no \\
PIRK \cite{pirk} with contraction bounds & yes & yes & no & no \\
HR \cite{hr} with hybridization & yes & no & yes & no \\
ProbReach \cite{probreach2} with $\delta$-reachability,  & no & no  & yes & no \\
VSPODE \cite{reliablecomput} using p-boxes & no &no  & yes & no \\
Gaussian process (GP) \cite{gp} & no & no & no & no \\
\textbf{LRT-NG \cite{gruenbacher2020lagrangian} (Ours)} & yes & no & yes & no \\
\textbf{Stochastic Lagrangian reachability SLR \cite{Gruenbacher2021verification} (Ours)} & no & yes & no & no \\
\textbf{GoTube\cite{Gruenbacher2021GoTube} (Ours)} & no & yes & no & \textbf{yes} \\
\bottomrule
\end{tabular}
\end{adjustbox}
 \label{tab:related_works}
\end{table*}
%
\section{Related Work}\label{sec:related work}
\subsubsection{Global Optimization.} Efficient local optimization methods such as gradient descent cannot be used for global optimization since optimization problems related to robustness analysis are typically non-convex. Thus, many advanced verification algorithms tend to use global optimization schemes \cite{NEURIPS2018_be53d253,bunel2020lagrangian}. Depending on the properties of the objective function, e.g., smoothness, various types of global optimization techniques exist. For instance, interval-based branch-and-bound (BaB) algorithms \cite{neumaier_2004,hansen2003global} work well on differentiable objectives up to a certain scale, which has recently been improved \cite{de2021improved}. There are also Lipschitz-global optimization methods for satisfying Lipschitz conditions \cite{Lipschitz,kvasov2013Lipschitz}. For example, a method for computing the Lipschitz constant of deep neural networks to assist with their robustness and verification analysis was recently proposed in~\cite{NEURIPS2019_95e1533e} and~\cite{bhowmick2021lipbab}. Additionally, there are evolutionary strategies for global optimization using the covariance matrix computation~\cite{cma,cmaes}. In our approach, for global optimization, we use random sampling and compute neighborhoods (Lipschitz caps) of the samples, where we have probabilistic knowledge about the values, such that we are able to correspondingly estimate the statistical global optimum with high confidence.~\cite{stochGlobOptim}.

\subsubsection{Verification of Neural Networks.}
A large body of work tried to enhance the robustness of neural networks against adversarial examples \cite{goodfellow2014explaining}. There are efforts that show how to break the many defense mechanisms proposed \cite{athalye2018obfuscated,lechner2021adversarial}, until the arrival of methods for formally verifying robustness to adversarial attacks around neighborhoods of data \cite{henzinger2021scalable}. 
The majority of these complete verification algorithms for neural networks work on piece-wise linear structures of small-to-medium-size feedforward networks \cite{NEURIPS2019_246a3c55}. For instance, \cite{bunel2020branch} has recently introduced a BaB method that outperforms state-of-the-art verification methods \cite{katz2017reluplex,NEURIPS2020_f6c2a0c4}. A more scalable approach for rectified linear unit (ReLU) networks \cite{nair2010rectified} was recently proposed based on Lagrangian decomposition; this approach significantly improves the speed and tightness of the bounds \cite{de2021improved}. The proposed approach not only improves the tightness of the bounds but also performs a novel branching that matches the performance of the learning-based methods \cite{lu2020nueral} and outperforms state-of-the-art methods \cite{NEURIPS2018_d04863f1,singheth,bak2020improved,henriksen2020efficient}. While these verification approaches work well for feedforward networks, they are not suitable for recurrent and continuous neural network instances, which we address. 

\subsubsection{Verification of Continuous-Time Systems.} Reachability analysis is a verification approach that provides safety guarantees for a given continuous dynamical system \cite{gurung2019parallel,vinod2021stochastic}. Most dynamical systems in safety-critical applications are highly nonlinear and uncertain in nature \cite{lechner2020neural}. The uncertainty can be in the system's parameters \cite{sreach,probreach,reliablecomput}, or their initial state~\cite{reliablecomput,10.1145/3126508}. This is often handled by considering balls of a certain radius around them. Nonlinearity might be inherent in the system dynamics or due to discrete mode-jumps \cite{e71eb65e23844408b72fe95a84f88cb6}. We provide a summary of methods developed for the reachability analysis of continuous-time ODEs in Table~\ref{tab:related_works}. 
A fundamental shortcoming of the majority of the methods described in Table~\ref{tab:related_works} is their lack of scalability while providing conservative bounds.
In this paper, we show that our algorithms establish the state-of-the-art for the verification of ODE-based systems in terms of speed, time-horizon, task completion, and scalability on a large set of experiments.

\section{Background}\label{sec:background}
\subsection{Reachability Analysis of ODEs}
For linear ordinary differential equations (ODEs) there exists a general closed-form solution, describing the behavior of the solution-traces over time, for every initial state. For nonlinear ODEs, there is no closed-form solution any more. One is able to calculate the solution for different initial states, but one does not know what happens in between of these already calculated traces. 

The main goal in the reachability analysis of nonlinear ODEs, is to over-approximate the reachable states of the ODEs, starting from a set of initial states, such as, an interval, a ball, or an ellipsoid, in a way that one can guarantee that all traces are inside the over-approximation. We call such an over-approximation a \textit{reachtube}. Let us now define this mathematically:
\begin{definition}[Initial value problem (IVP)]\label{def:IVP}
We have a time-invariant ordinary differential equation $\partial_t x = f(x), f:\RE^n\rightarrow\RE^n$, a set of initial values defined by a ball $\calB_0=B(x_0, \rd_x)$ with center $x_0\in\RE^n$ and radius $\rd_0\in\RE$, the initial condition $x(t_0) \in \calB_0$ and a sequence of $k$ timesteps $\{t_j\!:\!j\!\in\![1,\dots,k]\,\land\, (t_0 \!<\! t_1 \!<\! \cdots \!<\! t_k)\}$. For every $t_j$, we want to know the solution $x(t_j)$ of
\begin{align}\label{eq:IVP}
    \partial_t x = f(x),\quad x(t_0) \in \calB_0=B(x_0, \rd_x).
\end{align}
\end{definition}
The definition can be generalized to time-variant ODEs, as time can be just seen as an additional variable $x_{n+1}$ with $\partial_t x_{n+1} = 1$.
Let $\chij{x_0}=x(t_j)$ be the solution of Eq.~\eqref{eq:IVP} at time $t_j$, for $x(t_0)\,{=}\,x_0$. In reachability analysis, the goal is to find for every time step $t_j$ an overapproximation $\calB_j\,{\supseteq}\, \{\chij{x}:x\in\calB_0\}$, such that the set of these over-approximations build up a reachtube, containing the reachable states.
\begin{definition}[Reachtube]\label{def:Reachtube}
Given a set of initial values $\calB_0\,{\in}\,\RE^{n\times n}$, a nonlinear ODE as in Eq.~\eqref{eq:IVP}, the pointwise solution function $\chij{\cdot}\,{:}\,\RE^n\,{\rightarrow}\,\RE^n$ and over-approximations $\calB_j\,{\supseteq}\, \{\chij{x}\,{:}\,x\,{\in}\,\calB_0\}$. The Reachtube for a sequence of $k$ timesteps $\{t_j\,{:}\,j\,{\in}\,[1,\dots,k]\,{\land}\, t_0 < t_1 < \cdots < t_k)\}$ is defined as
\begin{align}
    \calR = \{\calB_0, \calB_1, \dots, \calB_k\}.
\end{align}
\end{definition}
As we are going to use balls and ellipsoids, we call $\calB_j$, \emph{bounding balls}. For every ellipsoid there is a metric such that the ellipsoid equals a ball in that metric. Let $M_j\,{\in}\,\RE^{n\times n}$ be a positive definite matrix ($M_j\,{\succ}\,0$), then there exists a decomposition: $A_j\,{\in}\,\RE^{n\times n}$ with $A_j^\top A_j\,{=}\,M_j$. Every ellipsoid can be defined as $B_{M_j}(x_j, \rd_j)\,{=}\,\{x:\|x\,{-}\,x_j\|_{M_j}\,{=}\,\rd_j\}$ with center $x_j$, weighted radius $\rd_j$ and norm $\|x\|_{M_j}\,{=}\,\sqrt{x^\top M_j x}\,{=}\,\|A_j x\|_2$. If $M_j$ is the identity matrix, then $\calB_j$ is a ball in the Euclidean metric, so we will omit the subscript and use $B(x_j, \rd_j)$.

When using reachability analysis to check for intersections with bad states, it is crucial to compute as tight as possible reachtubes.

\subsection{Interval Arithmetic and Lohner Method}
\begin{figure}
    \centering
    \includegraphics[width=0.6\textwidth]{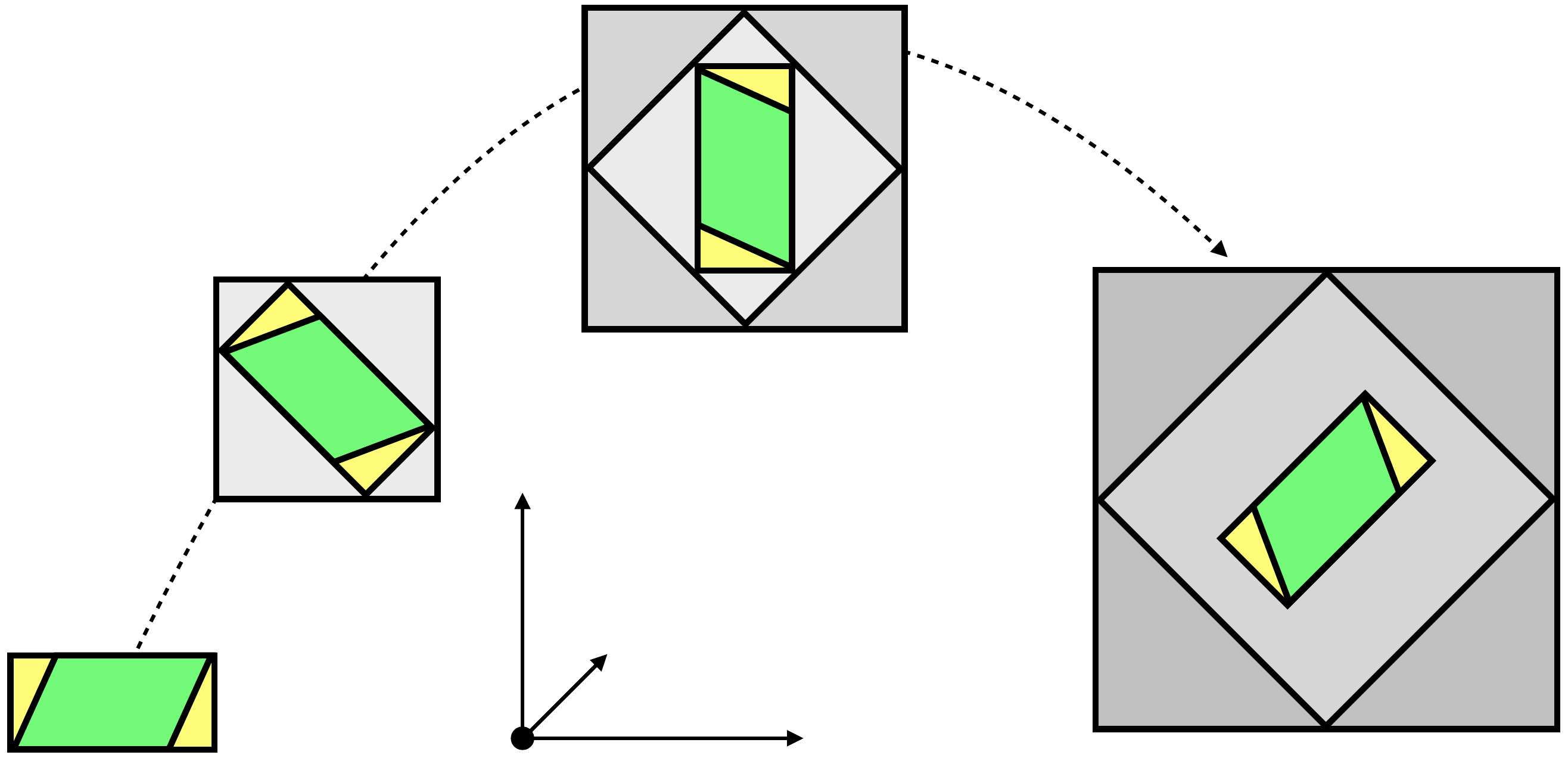}
    \caption{\emph{Wrapping effect.} Symbolic illustration for wrapping of a parallelogram (green) when applying a consecutive rotation of $45^\circ$ to it with interval boxes (grey) and with interval boxes in adapted coordinated systems using Lohner's QR method (yellow).}
    \label{fig:wrapping effect}
\end{figure}

There are different ways to define conservative regions by set representations: intervals, balls, ellipsoids, polytopes and more. In the papers \cite{gruenbacherArch19,gruenbacher2020lagrangian} we relied on interval arithmetic, so we want to shortly review the benefits and problems with that method.
The set of intervals on the real numbers is defined as (\cite{NEDIALKOV199921}):
\begin{align}
    \IR = \{[a]=[\underline{a},\overline{a}]:\underline{a}, \overline{a} \in \RE, \underline{a} \le \overline{a}\},
\end{align}
whereas an \emph{interval vector} $[x]\in\IR^n$ is a vector with interval components and an \emph{interval matrix} $[A]\in\IR^{n\times m}$ is a matrix with interval components.
The biggest problem in interval arithmetic is the wrapping effect, which happens if we apply concatenated functions on intervals (see Fig.~\ref{fig:wrapping effect}).
In the Lagrangian Reachability algorithms~\cite{gruenbacherArch19,gruenbacher2020lagrangian} an improved version of Lohner's QR method~\cite{Lo,NEDIALKOV199921} is used to directly address the wrapping effect caused by interval arithmetic. Intuitively, the rotational part $Q$ of a function evaluation is extracted, which is subsequently used as a new coordinate system. For a more detailed discussion of the steps mentioned above, please refer to~\cite{gruenbacherArch19}.

\subsection{Lipschitz Constant and the Variational Equation}
The Lipschitz constant defines a relation between the domain and the range of a function, more precisely it bounds the distance in the range by a multiple of the distance in the domain.
\begin{definition}[Local Lipschitz constant]\label{def:Lipschitz}
Let $\chi\! :A\!\rightarrow\!\RE^m$ ($A\!\subseteq \!\RE^n$) be a function, $M_A, M_B\!\succ\!0$ be respectively metrics on the domain and the range, $S\! \subseteq \! A$ be a subset of the domain and
\begin{align}
    \exists\lambda_S:\|\chi(x)-\chi(y)\|_{M_B} \le \lambda_S \|x-y\|_{M_A},\quad\forall x,y\in S,
\end{align}
then the smallest such $\lambda_S$ is called the Local Lipschitz constant of $\chi$ on set $S$.
\end{definition}
An upper bound of the Lipschitz constant can be computed using the mean value theorem from calculus with the statement either for scalar or for vector valued functions:
\begin{theorem}[Mean value theorem (generalized Rolle's Theorem)]\label{thm:mean value}
Let $M_1, M_2\!\succ\!0$ be respectively metrics on the domain and the range with $M_1 \!=\! A_1^\top A_1, M_2 \!=\! A_2^\top A_2$ and norm $\|x\|_{M_{1,2}}\! =\! \|A_2xA_1^{-1}\|_2$. Considering the change of metric~\cite[Lemma 2]{Cyranka2017} and the well-known mean value theorems, we are able to make the following statements:
\vspace*{-20pt}
\begin{enumerate}
\begin{align}
    \intertext{\item Let $d\! : A\! \rightarrow \! \RE$ ($A\!\subseteq \!\RE^n$) be a scalar function. Then it holds $\forall x,y\in A$:}
    \exists h\in[0,1]: \|d(x)-d(y)\|_{M_2} &\le \|\partial_x d(x + h\cdot (y-x))\|_{M_{1,2}} \cdot \|x-y\|_{M_1}
    \intertext{\item Let $\chi\! : A\! \rightarrow \! \RE^n$ ($A\!\subseteq \!\RE^n$) be a vector-valued function and the norm of the Jacobian matrix of $\chi$ be bounded by some constant $\Lambda_{1,2} \ge \|\partial_x \chi(x + h\cdot (y-x))\|_{M_{1,2}}$ for all $h\in[0,1]$ and all $x,y\in A$. Then it holds:}
    \|\chi(x)-\chi(y)\|_{M_2} &\le \Lambda_{1,2} \cdot \|x-y\|_{M_1}\quad\forall x,y\in A,
\end{align}
    and thus $\Lambda_{1,2}$ is an upper bound of the Lipschitz Constant $\lambda_S$ of Definition~\ref{def:Lipschitz}.
\end{enumerate}
\end{theorem}
The mean value theorem can be used to find an upper bound of the local Lipschitz constant. We will need such an upper bound for the deterministic as well as for the statistical guarantees. In both cases we will need to compute the jacobian matrix for the solution function $\chij{\cdot}$ of Eq.~\eqref{eq:IVP}, so the question is how to compute the jacobian matrix for the solution of a differential equation, for which we do not even have a closed form solution? For this purpose, we introduce $F_x\! :\! \RE \! \rightarrow \! \RE^{n\times n}$ with $F_x(t)=\partial_x\chit{x}$ called the \emph{deformation gradient} in~\cite{linTE,contMec}, and the \emph{sensitivity analysis} in~\cite{breach,donze}. $F_x(t)$ describes how much a small perturbation in the initial value $x$ changes the solution to the IVP at time $t$.
\begin{definition}[Variational equation]\label{def:variational equation}
Let $f$ be the system equations of the initial value problem defined in Eq.~\eqref{eq:IVP} and $\chit{x_0}$ be the solution at time $t$ for $x(t_0)=x_0$, then the following equation is called the variational equation:
\begin{align}\label{eq:variational equation}
    \partial_t F(t) = (\partial_x f)(\chit{x})F(t),\quad F(t_0)=I,
\end{align}
with $I \in \RE^{n\times n}$ being the identity matrix.
\end{definition}
Intuitively, Def.~\ref{def:variational equation} describes how an initial perturbation in the initial value evolves over time. In~\cite{gruenbacher2020lagrangian} it was shown that $F_x$ is a solution of the \emph{variational equations} associated to the system equations in Eq.~\eqref{eq:IVP}.

\subsection{Continuous-Depth Models}
ODE's are used to describe the dynamics of the hidden states of continuous-depth neural models~\cite{neuralODEs}. The output is a function of the solution of the ODE at a given time. So the derivative of the unknown states $x$ is described by a parameterized vector-valued function $f_\theta\! : \! \RE^{n}\! \rightarrow \! \RE^n$, which is assumed to be Lipschitz-continuous and forward-complete:
\begin{align}\label{eq:neuralODE}
    \partial_t x = f_\theta(x),\quad x(t_0)\in\calB_0
\end{align}
By adding time as an additional variable $x_{n+1}$ with $\partial_t x_{n+1} = 1$, a continuous-depth model can be seen as a special case of Eq.~\eqref{eq:IVP}.
\section{Deterministic Guarantees (LRT-NG)}\label{sec:LRT-NG}
The most straightforward way to compute a conservative reachtube as defined in Def.~\ref{def:Reachtube}, would be to use an interval enclosure $[\calX_0]\!\supseteq\! \calB_0$ of the initial values and just use interval-arithmetic evaluations of an integration method, for example, the Runge-Kutta method, to propagate them from timestep to timestep. 

Due to the infamous wrapping effect (as shown in Fig.~\ref{fig:wrapping effect}), this would lead very soon to a blow-up in space. As already mentioned in the related work in Section~\ref{sec:related work}, there are different approaches on how to avoid that blow-up in space and create as tight as possible reachtubes.

Lagrangian Reachability is a bloating based technique: starting with an initial ball $\calB_0\!=\!B(x_0,\rd_0)$, at a sequence of $k$ timesteps $\{t_j\!:\!j\!\in\![1,\dots,k]\,\land\, (t_0 \!<\! t_1 \!<\! \cdots \!<\! t_k)\}$, it propagates the center of the ball and computes the new radius $\rd_j$, by multiplying $\rd_0$ with a stretching factor $\Lambda_j$.
\begin{algorithm}[t]
    \caption{LRT-NG}
    \label{algorithm:LRT-NG}
    \begin{algorithmic}[1]
    \REQUIRE initial ball $\calB_0\! =\! B(x_0, \rd_0)$, initial metric $M_0$, initial metric decomposition $A_0~(M_0\! = A_0^\top A_0)$, time horizon T, sequence of timesteps $t_j$ ($t_0\!<\!\dots\!<\! t_k=T$), system dynamics $f$
    \vspace*{2mm}
    \STATE \textbf{set} $\quad [\calF] \leftarrow \{I\}, [\calX] \leftarrow$ overapproximation of $\calB_0$
    \FOR{$(j=1; j\le k; j=j+1)$}
        \STATE $x_j \leftarrow$ \emph{solveIVP}($f, x_{j-1}, [t_{j-1},t_j]$)
        \STATE $[\calF] \leftarrow F_{[\calX_0]}(t_j) = $ \textit{rungeKuttaVariational}$((\partial_x f)([\calX]),[\calF],[t_{j-1},t_j]))$
        \STATE $M_j \leftarrow$ \textit{computeOptimalMetric}$(F_{x_j}(t_j), A_0)$\label{line:metric computation}
        \FORALL{$M \in \{M_j, I\}$}
            \STATE \textbf{compute} $\Lambda \ge \|[\calF]\|_{M}$ (stretching factor)
        \ENDFOR
        \STATE $\calB_j \leftarrow B_{M_j}(x_j,\rd_{M_j})$
        \STATE $\calB_j^{circle} \leftarrow B(x_j,\rd_{I})$
        \STATE $[\calX] \leftarrow$ \textit{intersectionBox}($\calB_{j}, \calB_j^{circle}$)\label{line:intersection box}
    \ENDFOR
    \RETURN $(\calB_1,\dots,\calB_k), (\calB_1^{circle},\dots,\calB_k^{circle})$
    \end{algorithmic}
\end{algorithm}
Using Thm.~\ref{thm:mean value} it holds that:
\begin{align}\label{eq:mean value}
\begin{split}
\max_{x\in\calB_0}\lVert\chij{x}-\chij{x_0}\rVert_{M_j}
&\leq \max_{x\in\calB_0}\lVert F_x(t_j)\rVert_{M_{j}} \max_{x\in\calB_0}\lVert x-x_0\rVert_{M_0}\\
\end{split}
\end{align}
We compute $\max_{x\in\calB_0}\lVert F_x(t_j)\rVert_{M_j}$ by using interval arithmetic to propagate all possible deformation gradients as an interval $[\calF_j]\supseteq\{F_x(t):x\in\calB_0\}$ with an interval arithmetic version of the variational equation Eq.~\eqref{eq:variational equation}:
\begin{align}\label{eq:interval variational equation}
    \partial_t [\calF] = (\partial_x f)([\calX_t])[\calF],
\end{align}
where $[\calX_{t_j}]$ is an as-tight-as-possible interval overapproximation of $\calB_j$.
Thus the challenge is to bound the norm of the interval deformation gradients:
\begin{align}\label{eq:Lambda}
    \lVert [\calF_j]\rVert_{M_{j}}\leq \Lambda_{j} \Rightarrow \rd_j = \Lambda_{j}\rd_0.
\end{align}
We call $\Lambda_{j}$ - the upper bound of the Lipschitz constant of $\chij{\cdot}$ - the \emph{stretching factor (SF)} associated to the interval gradient tensor, as it shows by how much the initial ball $\calB_0$ has to be stretched, such that it encloses the set of all reachable states.
Having the interval gradient $[\calF_j]$ at time $t_j$ we solve Equation~\eqref{eq:Lambda} using algorithms from~\cite{hladik,rump,rohn}, and choosing the tightest result available. The correctness of Lagrangian Reachability is rooted in~\cite[Theorem 1]{Cyranka2017}.

As the tightness of the bounding balls $\calB_j$ depends on the previous values, for example $\calB_{j-1}, [\calX_j]$ or $[\calF_j]$, the wrapping deficiencies accumulate in time, as shown in Fig.~\ref{fig:wrapping effect}. This is why the theoretical advances of LRT-NG concentrate on minimizing the volume of the bounding balls and their enclosure, and thus on creating tighter and longer reachtubes.
\subsection{Lagrangian Reachtubes: The Next Generation}\label{sec:LRTNG}
This section presents the theoretical advances of LRT-NG in~\cite{gruenbacher2020lagrangian}. In particular, we first state the optimization problem to be solved in order to get the optimal metric, and thus the bounding ball with minimal volume. We first describe an analytic solution of an optimal metric minimizing the volume of the ellipsoid and prove that it solves the optimization. Finally, we focus on the new reachset box $[\calX_j]$ computation, the interval overapproximation of the ellipsoid-ball-intersection.

As shown in Algorithm~\ref{algorithm:LRT-NG}, LRT-NG iterates over the sequence of $k$ timesteps, until it reaches the given time horizon $T$. After propagating the center point, it computes the interval deformation gradient by integrating Eq.~\eqref{eq:interval variational equation} in line~\ref{line:variational equation}. After computing the optimal metric $M_j$, it bounds the maximum singular value of $[\calF]$ in the Euclidean norm, as well as in $M_j$ norm, such that LRT-NG constructs an ellipsoid $\calB_j$ and Euclidean bounding ball $\calB_j^{circle}$. This enables us to define an as-tight-as-possible interval box $[\calX]$, over the intersection of the ellipsoid and the ball. This intersection-based approach considerably reduces the wrapping effect of the next integration of the interval variational equation.
\subsubsection{Computation of the Metric}\label{sec:metric}
To obtain an as-tight-as-possible over-approxi\-ma\-tion, we wish to minimize the volume of the n-dimensional ball $\calB_j$, that is, of $B_{M_j}(x_j,\delta_j)$. Hence, the optimization problem is given by:
\begin{align}
    \argmin_{M_j\succ 0} \vol (B_{M_j}(x_j,\delta_j)),
\end{align}
where $\delta_j\!=\!\Lambda_j(M_j)\!\cdot\!\rd_0$. Let us further define $\hat{F}_{j-1,j}\!=\! \partial_x\chi_{t_{j-1}}^{t_j}(x)|_{x=x_{j-1}}$ as the deformation gradient from time $t_{j-1}$ to $t_j$ at the center of the ball. Using the chain rule it holds that $F_j \!=\! \prod_{m=1}^j \hat{F}_{m-1,m}$, where $F_j$ is defined as the deformation gradient at $x_0$.
\begin{figure}[t]
    \centering
    \includegraphics[width=0.6\textwidth]{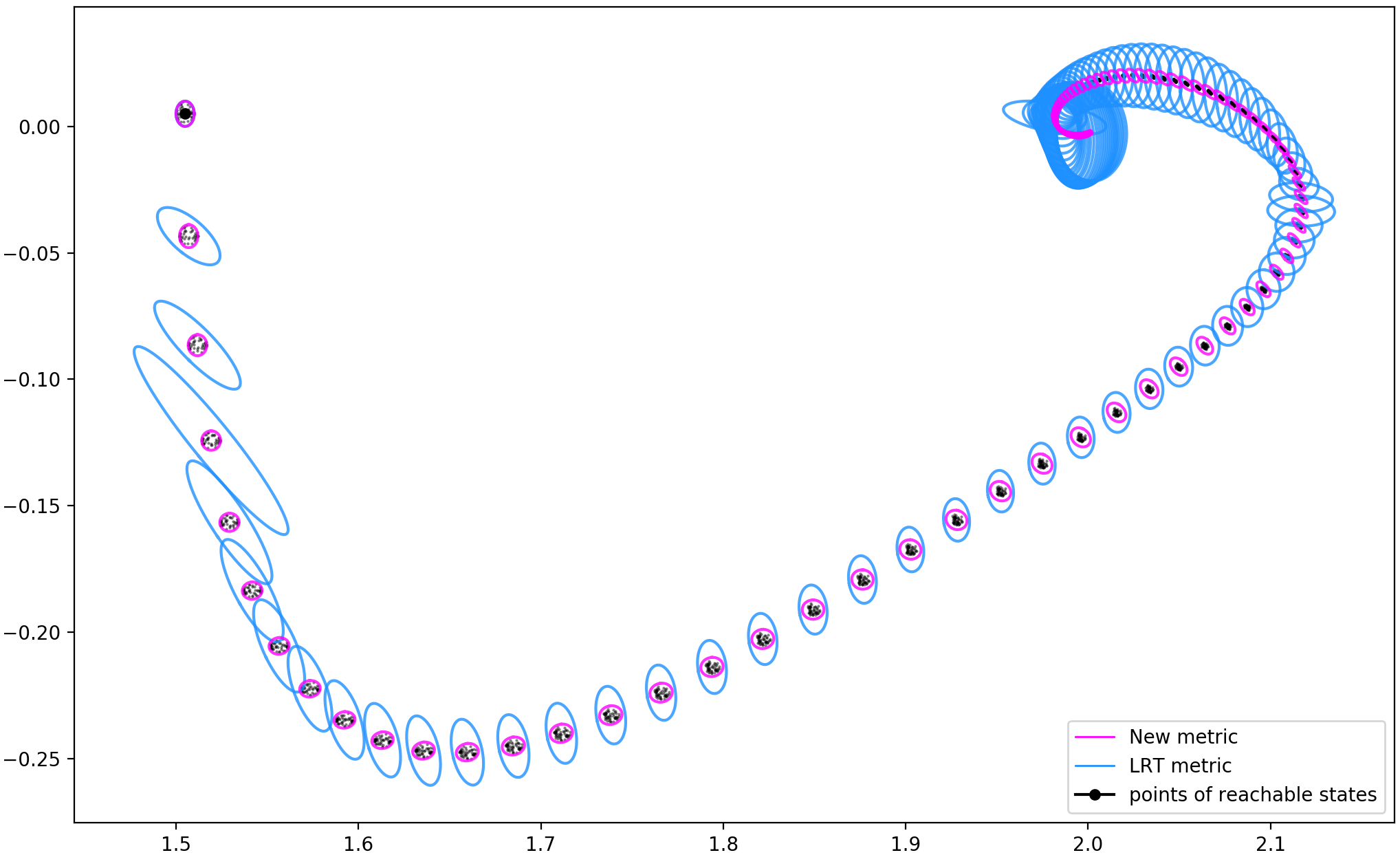}
    \vspace{-1.ex}
    \caption{Reachtube for the Robotarm model, obtained with the LRT (in blue) and LRT-NG metric (in purple), respectively. The time is bounded to the interval $t\in[0,5]$, and it evolves starting at the top left corner of the figure, and going to the right.}
    \label{fig:metricComparison}
\vspace{-3ex}
\end{figure}
The following theorem defines a metric $\hat{M}_j$ and shows that this metric minimizes the ellipsoid volume, and it is therefore optimal.
\begin{theorem}[Thm. 1 in~\cite{gruenbacher2020lagrangian}]\label{thmopt}
Let the gradient-of-the-flow matrices $F_j$ and $\hat{F}_{j-1,j}\,{\in}\,\R^{n\times n}$ be full rank, and the coordinate-system matrix of the last time-step $A_{j-1}\,{\in}\,\R^{n\times n}$ be full-rank and $A_{j-1}\,{\succ}\,0$. Define metric $\hat{M}_j(F_{j})\!=\!\hat{A}_j(F_j)^\top\hat{A}_j(F_j)$:
\begin{align}\label{M1opt}
    \hat{A}_j(F_j) = A_{j-1}\hat{F}_{j-1,j}^{-1} = A_0F_j^{-1}
\end{align}
When $F_j$ is known, we simply abbreviate $\hat{A}_j(F_j)$ with $\hat{A}_j$, and $\hatM_j(F_j)$ with $\hatM_j$. Let $\Lambda_{0,j}(M_j)$ be given by (with $M_0$ fixed):
\[
\Lambda_{0,j}(M_j)=\sqrt{\lambda_{\max}\left((A_0^\top)^{-1}F_j^\top M_j F_j A_0^{-1}\right)}.
\]
Then, it holds that $\ \vol\left(B_{\hat{M_j}}(\chi_{t_0}^{t_j}(x_0),\Lambda_{0,j}(\hat{M_j})\,\delta_0)\right)$ is equal to:
\[
    \min_{M_j\succ 0} \vol\left(B_{M_j}(\chi_{t_0}^{t_j}(x_0),\Lambda_{0,j}(M_j)\,\delta_0)\right).
\]
In other words, the symmetric matrix $\hatM_j\,{\succ}\,0$ minimizes the volume of the ellipsoid $B_{M_j}(\chi_{t_0}^{t_j}(x_0),\Lambda_{0,j}(M_j)\,\delta_0)$ as a function of $M_j$.
\end{theorem}
Thus, Theorem~\ref{thmopt} gives us an analytic solution for the optimal metric, releasing us from either solving an optimization problem with semi-definite programming in every time-step like in~\cite{fansimul,Cyranka2017}, or risking false-positive consequences of using a suboptimal metric as in LRT~\cite{CyrankaCDC18,gruenbacherArch19}.

A comparison of the reachsets obtained with LRT metric $\tilde{M}_j$~\cite[Definition 1]{CyrankaCDC18} and the one obtained with LRT-NG metric $\hat{M}_j$ is illustrated in Fig.~\ref{fig:metricComparison}. It shows that the LRT metric is by far not an optimal choice, and it also shows how well our new analytically computed metric $\hat{M}_j$ follows the shape of the set of reachable states.
\subsubsection{Intersection of the Bounding Balls}\label{sec:reacxh}
Another novelty in LRT-NG, is that the next reachset is the intersection of an ellipsoid computed in the optimal metric and an Euclidean ball. This considerably reduces the volume and therefore enables LRT-NG to work also for continuous-depth models.

An effective way of getting a much tighter conservative bound $[\calX_j]$ is taking the intersection of the ellipsoid in the optimal metric $\hatM_j$, and the ball in Euclidean metric. As small errors accumulate in interval arithmetic, taking the intersection leads to a considerable improvement especially as the time horizon increases. That new approach is conservative is shown in Lemma 1 of~\cite{gruenbacher2020lagrangian}, which allows us to dramatically reduce the volume of the reachtube and combat the wrapping effect in a way that has not been considered before by bloating-based techniques~\cite{Cyranka2017,CyrankaCDC18,gruenbacherArch19,fansimul,Fan2016,Fan2015}.
\section{Statistical Guarantees (SLR \& GoTube)}\label{sec:SLR and GoTube}
To avoid state explosion as in the conservative methods, we developed a statistical version of Lagrangian reachability and provided convergence guarantees for computing the upper bound of the confidence interval for the maximum perturbation at time $t_j$ with confidence level $1\,{-}\,\gamma$ and tube tightness $\mu$. 

We review first SLR, a purely theoretical statistical version of Lagrangian reachability framework~\cite{Gruenbacher2021verification}, and then GoTube, a practical statistical verification algorithm for continuous-time models~\cite{Gruenbacher2021GoTube}, where we achieved technical solutions for fundamental issues in applying SLR. 

In this work, we describe \emph{reachability as an optimization problem} and solve that problem for every timestep such that the size of the bounding ball $\calB_j$ at time $t_j$ does not depend on the previous values $\calB_{j-1}, [\calX_j]$ or $[\calF_j]$ like in LRT-NG.
To compute a bounding tube, we have to compute at every time step $t_j$, the maximum perturbation $\rd_j$ in metric $M_j$ for $x\,{\in}\,\calB_0$, which is defined as a solution of the optimization problem:
\begin{align}\label{eq:optim}
    \rd_j \ge \max_{x\in\calB_0}\|\chi(t_j, x) - \chi(t_j, x_0)\|_{M_j} = \max_{x\in\calB_0} d(\chij{x}) = m^\star,
\end{align}
where $d_j(x)=d(\chij{x})$ denotes the {\em distance} at time $t_j$, if the initial center $x_0$ and metric $M_j$ is known from the context.
\begin{figure}[t]
\centering
\includegraphics[width=0.8\textwidth]{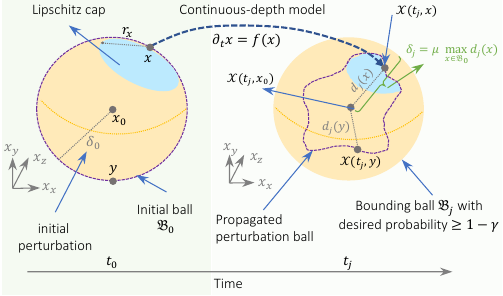}
\caption{Statistical Guarantees in a nutshell. The center $x_0$ of ball $\calB_0\,{=}\,B(x_0, \rd_0)$, with $\rd_0$ the initial perturbation, and samples $x$ drawn uniformly from $\calB_0$'s surface, are numerically integrated in time to $\chi(t_j, x_0)$ and $\chi(t_j, x)$, respectively. The Lipschitz constant of $\chi(t_j, x)$ and their distance $d_j(x)$ to $\chi(t_j, x_0)$ are then used to compute Lipschitz caps around samples $x$, and the radius $\rd_j$ of bounding ball $\calB_j$ depending on the chosen tightness factor $\mu$. The ratio between the caps' surfaces and $\calB_0$'s surface are correlated to the desired confidence $1\,{-}\,\gamma$. }
\label{fig:gotube}
\vspace*{-4ex}
\end{figure}

As we require Lipschitz-continuity and forward-completeness of the con\-tinuous-depth model in Eq.~\eqref{eq:neuralODE}, the map $x\! \mapsto\! \chij{x}$ is a homeomorphism and commutes with closure and interior operators. In particular, the image of the boundary of the set $\calB_0$ is equal to the boundary of the image $\chij{\calB_0}$. 
Thus, Eq.~\eqref{eq:optim} has its optimum on the surface of the initial ball $\calB_0^S = \textrm{surface}(\calB_0)$, and we will only consider points on the surface. 

In order to be able to optimize this problem, we describe the points on the surface with (n-dimensional) polar coordinates such that every point $x\,{\in}\,\calB_0^S$ is represented by a tuple $(\rd_0,\varphi)$, with angles $\varphi \,{=}\, (\varphi_1,\dots,\varphi_{n-1})$ and center $x_0$, having a conversion function $x((\rd_0,\varphi),x_0)$ from polar to Cartesian coordinates.
Whenever the center $x_0$ and the radius $\rd_0$ of the initial ball $\calB_0$ are known from the context, we will use the following notation: $x(\varphi)$ for the conversion from polar to Cartesian coordinates, and just $x$ if we do not want to mention the polar coordinates explicitly.
\subsubsection{Forward-Mode Use of Adjoint-Sensitivity Method.} For both algorithms, we will need $F_x$ for several sample points. The integral of Eq.~\eqref{eq:variational equation} has the same form as the auxiliary ODE used for reverse-mode automatic differentiation of Neural ODEs, when optimized by the adjoint sensitivity method \cite{neuralODEs} with one exception: our $F_x$ defines the differential of the solution by the initial value and their equivalent function $a$ defines the differential of the loss function by the initial value. Their approach computes gradients by solving a second, augmented ODE backwards in time. In our case, solving the variational Eq.~\eqref{eq:variational equation} until target time $t_j$, already gives us the required gradient $F_x$, but requires knowledge of $\chit x$ for all $t\in[t_0,t_j]$. This is why in our forward-mode adjoint sensitivity method, we propagate for all samples $x$ using Eq.~\eqref{eq:IVP} and $F_x(t)$ using Eq.~\eqref{eq:variational equation} forwards in time together until $t_j$, starting from its augmented (combined) initial state $(x,I)$ and using its augmented dynamical system $(f(x), (\partial_x f)(\chit{x})F(t))$.
\subsection{Theoretical Statistical Verification Framework}\label{sec:SLR}
In this section, we review \emph{stochastic Lagrangian reachability} (SLR) from~\cite{Gruenbacher2021verification}, where we solve each optimization problem globally, via uniform sampling, and locally, through gradient descent, whereas gradient descent is avoided in spherical-caps around the start/end states of previous searches. The cap radius is derived from its local Lipschitz constant, computed via interval arithmetic.
\begin{algorithm}[t]
    \caption{Stochastic Lagrangian Reachability (SLR)}
    \label{algorithm:SLR}
    \begin{algorithmic}[1]
    \REQUIRE initial ball $\calB_0\!=\!B(x_0,\rd_0)$, time horizon T, sequence of timesteps $t_j$ ($t_0\le t_1\le\dots\le t_k=T$), tolerance $\mu\,{>}\,1$, confidence level $\gamma\,{\in}\, (0,1)$, distance function $d_j$, gradient of loss $\nabla_\varphi L$
    \vspace*{2mm}
    \FOR{$(j=1; j\le k; j=j+1)$}
        \STATE $\calV, \calU \leftarrow \{\}$ \quad(list of visited and random points)
        \STATE $x_j \leftarrow$ \emph{solveIVP}($f, x_{j-1}, [t_{j-1},t_j]$)
        \STATE $[\calF] \leftarrow F_{[\calX_0]}(t_j) = $ \textit{rungeKuttaVariational}$((\partial_x f)([\calX]),[\calF],[t_{j-1},t_j]))$\label{line:variational equation}
        \STATE \textbf{compute} $\Lambda \ge \|[\calF]\|$ (interval arithmetic Lipschitz constant)
        \STATE $\calP\leftarrow 0$
        \WHILE{$\calP < 1 - \gamma$}
            \STATE \textbf{sample} $x \in \calB_0$ and add sample to $\calV$ and $\calU$
            \STATE $\chij{x}, F_{x}(t_j) \leftarrow$ \emph{forwardModeAdjointSensitivity}($x, I, [t_0,t_j]$)
            \STATE \textbf{if} $x \notin \calS$ \textbf{then} \emph{findLocalMinimum}($x,\nabla_\varphi L,F_{x}(t_j)$)\label{line:gradient descent} and add $x$ to $\calV$
            \STATE $\bar{m}\leftarrow \max_{x\in\calV} d_j(x)$
            \STATE $r_x \leftarrow$ \emph{computeSafetyRegionRadius}($d_j(x), \bar{m}, \Lambda$) $\quad\forall x\in\calV$
            \STATE $\calS\leftarrow\bigcup_{x\in\calV} B(x,r_x)$
            \STATE $\calP \leftarrow \Pr(\mu \cdot \bar{m} \le m^\star)$\label{line:SLR probability}
        \ENDWHILE
        \STATE $\calB_j\leftarrow B(x_j, \mu\cdot\bar{m})$
    \ENDFOR
    \RETURN $(\calB_1,\dots,\calB_k)$
    \end{algorithmic}
\end{algorithm}
%
\subsubsection{Gradient Computation}\label{sec:gradient computation}
The SLR algorithm uses gradient descent locally, when solving the global optimization problem of Eq.~\eqref{eq:optim}. In~\cite{Gruenbacher2021verification} the \emph{loss function} $L(\varphi)\! = - d_j\! \circ x(\varphi)$ is introduced in polar coordinates at time $t_j$ to be able to do gradient descent on the surface in order to find the optimum. Note that $\Lp$ also depends on the initial radius $\delta _0$ and initial center $x_0$; as these are fixed inputs, we do not consider them in the notation. Gradient descent is started from uniformly sampled points, which are not contained in already constructed safety regions.

In~\cite{Gruenbacher2021verification}, we introduced a new framework to compute the loss's gradient which is needed to find the local minimum ain a unified fashionnd improved the optimization runtime by 50\%, compared to the optimization scheme used in \cite{neuralODEs}: we save half of the time because we do not have to go backward to compute the loss. 
\subsubsection{Safety-Region Computation}\label{sec:TR and TD}
In contrast to~\cite{Gruenbacher2021verification}, we will switch back to talking about a global maximum of $d_j(x)$ for points $x\in\calB_0$ instead of the equivalent problem of finding a global minimum of $L(\varphi)$ for points $\varphi\in\RE^{n-1}$. With our global search strategy, we are covering the feasible region $\calB_0^S$ with already visited points $\calV$. Consequently, we have access to the current maximum in $\calV$:
\begin{align}\label{eq:local minimum}
    \bar{m}_{j,\calV} = \max_{x\in\calV}d_j(x)
\end{align}
with $\bar{m}\le m^\star$, where $m^\star$ is the global maximum of Eq.~\eqref{eq:optim}. We now identify safety regions for a continuous-depth model flow and describe how this is incorporated in the SLR algorithm.  
\begin{definition}[Safety Region]\label{def:TR}
    Let $x_i\,{\in}\,\calV\subseteq\calB_0$ be an already-visited point. A safety-radius $r_{x_i}\,{=}\,r(x_i)$ defines a \emph{safe spherical-cap} $B(x_i,r_{x_i})^S \,{=}\, B(x_i,r_{x_i}) \cap \calB_0^S$, if it holds that $d_j(y)\le\mu\cdot\bar{m}$ for all $y\!\in B(x_i,r_x)^S$.
\end{definition}
In the following theorem, we use Thm.~\ref{thm:mean value} to bound the local Lipschitz constant (Def.~\ref{def:Lipschitz}) and to define the radius $r_x$ of the safety region $ B(x,r_x)^S$ around an already-visited point $x\in\calV$.
\begin{theorem}[Radius of Safety Region (\cite{Gruenbacher2021verification}, Thm. 1)]\label{thm:safety region radius}
    At target time $t_j$, let $\bar{m}$ be the current global maximum, as in Eq.~\eqref{eq:local minimum}.
    Let $x\!\in\!\calV$ be an already-visited point with value $d_j(x)$, and let $r_x$ and $ B(x,r_x)^S$ be defined as follows:
    \begin{align}\label{eq:safety radius}
        r_{x} =
        \lambda_{\Sigma_x}^{-1}\left(\mu\cdot\bar{m} - d_j(x)\right)
    \end{align}
    with $\mu > 1$, $\lambda_{\Sigma_x} =
    \max_{y\in\Sigma_x}\lVert F_y(t_j) \rVert_{M_{0,j}}$ and $\Sigma_x\supseteq  B(x,r_x)^S$, then it holds that:
    \begin{align}\label{eq:safety radius result}
        d_j(y)\le \mu\cdot\bar{m}\quad\forall y\in  B(x,r_x)^S
    \end{align}
\end{theorem}
We can now use the safety regions around the samples to compute the probability needed in Line~\ref{line:SLR probability} of Algorithm~\ref{algorithm:SLR}:
\begin{align}
    \Pr(\mu\cdot\bar{m} \ge m^\star) \ge \Pr(\exists x\in\calU: S_x\owns x^\star)=1 - \prod_{x\in\calU} \left(
    1 - \Pr(S_x)
    \right),
\end{align}
with $S_x\!=\!B(x,r_x)^S$ being the safety region around $x$. In~\cite{Gruenbacher2021verification}, we provide a convergence guarantee as well as a convergence rate for the probability $\Pr(S_x)\!=\!\area(S_x) / \area(\calB_0)$. Theorem 2 of~\cite{Gruenbacher2021verification} shows that in the limit of the number of samples, the constructed reachset converges with probability~1 to the smallest ellipsoid that encloses the true reachable set using tightness bound $\mu$.

\subsection{Scalable Statistical Verification}\label{sec:GoTube}
As we implemented the SLR algorithm, we observed that even after resolving the first-occurring inefficient sampling and their vanishing gradient problems, the algorithm still blew up in time, even for low-dimensional benchmarks such as the Dubins Car. Our GoTube algorithm and its associated theory solve fundamental scalability problems of related works (see Table~\ref{tab:related_works}) by replacing the interval arithmetic used to compute deterministic caps, with statistical Lipschitz caps. This enables us to verify continuous-depth models up to an arbitrary time-horizon, a capability beyond what was achievable before.


\begin{algorithm}[t]
    \caption{GoTube}
    \label{algorithm:GoTube}
    \begin{algorithmic}[1]
    \REQUIRE initial ball $\calB_0 = B(x_0, \rd_0)$, time horizon T, sequence of timesteps $t_j$ ($t_0<\dots< t_k=T$), error tolerance $\mu\,{>}\,1$, confidence level $\gamma\,{\in}\, (0,1)$, batch size $b$, distance function $d$
    \vspace*{2mm}
    \STATE $\calV\leftarrow\{\}$ \quad(list of visited random points)
    \STATE \textbf{sample batch} $x^B \in \calB_0^S$
    \FOR{$(j=1; j\le k; j=j+1)$}
        \STATE $\calP\leftarrow 0$
        \WHILE{$\calP < 1 - \gamma$}
            \STATE $\calV\leftarrow \calV \cup \{x^B\}$
            \STATE $x_j \leftarrow$ \emph{solveIVP}($f, x_{j-1}, [t_{j-1},t_j]$)
            \STATE $\bar{m}_{j,\calV} \leftarrow \max_{x\in\calV} d(t_j, x)$
            \STATE $x, F_x(t_j) \leftarrow$ \emph{forwardModeAdjointSensitivity}($x, I, [t_0,t_j])\quad\forall x\in\calV$
            \STATE \textbf{compute} local Lipschitz constants $\lambda_x = \|F_x\|$ for $x\in\calV$
            \STATE \textbf{compute} statistical quantile $\Delta\lambda_{\calV}$
            \STATE \textbf{compute} cap radii $r_x(\lambda_x, \Delta\lambda_{\calV})$ (Thm.~\ref{thm:Lipschitz cap}) for $x\in\calV$
            \STATE $\calP\,\leftarrow\, \textit{computeProb}(\gamma, \{r_x: x \in \mathcal{V}\}, n, \delta_0)$\label{line:computeProb}
            \STATE \textbf{sample batch} $x^B \in \calB_0$\label{line:new batch}
        \ENDWHILE
        \STATE $\calB_j\leftarrow B(x_j, \mu\cdot\bar{m}_{j,\calV})$
    \ENDFOR
    \RETURN $(\calB_1,\dots,\calB_k)$
    \end{algorithmic}
\end{algorithm}


GoTube starts by sampling a batch (tensor) $x^B\in\calB_0^S$ and if needed, it adds new samples to that tensor in Line~\ref{line:new batch} and computes every step in a tensorized manner, for all samples at the same time. In each iteration, it integrates the center and the already available samples from their previous time step, and the possibly new batches from their initial state (for simplicity, the pseudocode does not make this distinction explicit). GoTube then computes the maximum distance from the integrated samples to the integrated center, their local Lipschitz constant according to the variational Equation~\eqref{eq:variational equation} using the forward-mode adjoint sensitivity method. Unlike SLR, $F_x$ is not used with gradient descent to find local optima, but to compute local Lipschitz constants for all samples. 

Based on this information GoTube then computes a statistical upper bound for Lipschitz constants and the cap radii accordingly. The total surface of the caps is then employed to compute and update the achieved confidence (that is, probability). Once the desired confidence is achieved, GoTube exits the inner loop, and computes the bounding ball in terms of its center and radius, which is given by tightness factor $\mu$ times the maximum distance $\bar{m}_{j,\calV}$. After exiting the outer loop, GoTube returns the bounding tube.
\begin{definition}[Lipschitz Cap]
    Let $\calV$ be the set of all sampled points, $x\in\calV$ be a sample point on the surface of the initial ball, $\bar{m}_{j,\calV} = \max_{x\in\calV} d_j(x)$ be the sample maximum, and $B(x,r_x)^S = B(x,r_x)\cap\calB_0^S$ be a spherical cap around that point. We call the cap $B(x,r_x)^S$ a $\gamma,t_j$-Lipschitz cap if it holds that $\Pr\left(d_j(y) \le \mu\cdot \bar{m}_{j,\calV}\right)\ge 1-\gamma$ for all $y\in B(x,r_x)^S$.
\end{definition}
Lipschitz caps around the samples, are a statistical version of the safety regions around samples, commonly used to cover a state space. Intuitively, the points within a cap do not have to be explored. The difference with Lipschitz caps is that we statistically bound the values inside that space, and develop a theory enabling us to calculate a probability of having found an upper bound of the true maximum $m_j^\star = d_j(x_j^\star) = \max_{\{x_1,\dots,x_m\}\subset\calB_0} d_j(x)$ for any $m$-dimensional set of the optimization problem in Eq.~\eqref{eq:optim}.

Our objective is to avoid the usage of interval arithmetic, for computing the Lipschitz constant - as done in SLR - as it impedes scaling up to continuous depth models. Instead, we define statistical bounds on the Lipschitz constant, to set the radius $r_x$ of the Lipschitz caps, such that $\mu\cdot\bar{m}_{j,\calV}$ is a $\gamma$-statistical upper bound for all distances $d_j(y)$ at time $t_j$, from values inside the ball $B(x, r_x)^S$.
\begin{theorem}[Radius of Statistical Lipschitz Caps (\cite{Gruenbacher2021GoTube}, Thm. 1)]\label{thm:Lipschitz cap}
    Given a continuous-depth model $f$ from Eq.~\eqref{eq:IVP}, $\gamma\,{\in}\,(0,1)$, $\mu\,{>}\,1$, target time $t_j$,
    the set of all sampled points $\calV$, the number of sampled points $N\,{=}\,|\calV|$, the sample maximum $\bar{m}_{j,\calV}\,{=}\,\max_{x\in\calV} d_j(x)$, the IVP solutions $\chi(t_j,x)$, and the corresponding stretching factors $\lambda_x\,{=}\,\|\partial_x\chi(t_j,x)\|$ for all $x\,{\in}\,\calV$, then: let $\nu_x\,{=}\,|\lambda_x\,{-}\,\lambda_X|/\|x\,{-}\,X\|$, for $x\,{\in}\,\calV$, be a new random variable, where $X\,{\in}\,\calB_0^S$ is the random variable which is thrown by random sampling on the surface of the initial ball. Let the upper bound $\Delta\lambda_{\calV}$ of the confidence interval of $\E\nu_x$ be defined as follows:
    \begin{align}\label{eq:expected difference quotient}
        \Delta\lambda_{\calV}(\gamma) = \overline{\nu_x} + t^*_{\gamma/2}(N-2) \frac{s(\nu_x)}{\sqrt{N-1}},
    \end{align}
    with $\overline{\nu_x}$ and $s(\nu_x)$ being the sample mean and sample standard deviation of $\nu_x$, and $t^*$ being the Student's $t$-distribution. Let $r_x$ be defined as:
    \begin{align}\label{eq:cap radius}
        r_{x} = \frac{\left(-\lambda_x + \sqrt{\lambda_x^2 + 4\cdot\Delta\lambda_{\calV}\cdot(\mu\cdot\bar{m}_{j,\calV}-d_j(x))}\right)}{2\cdot\Delta\lambda_{\calV}},
    \end{align}
    then it holds that:
    \begin{align}\label{eq:cap probability}
        \Pr\left(d_j(y) \le \mu\cdot \bar{m}_{j,\calV}\right)\ge 1-\gamma\quad \forall y\in B(x,r_x)^S,
    \end{align}
    and thus that $B(x, r_x)^S$ is a $\gamma, t_j$-Lipschitz cap.
\end{theorem}
Using conditional probabilities, we are able to state that the convergence guarantee holds for the GoTube algorithm, thus ensuring that the algorithm terminates in finite time, even using statistical Lipschitz caps around the samples, instead of deterministic local balls~\cite{Gruenbacher2021GoTube}[Thm. 2].

\section{Experimental Evaluation}\label{sec:experiments}
We performed a rich set of experiments with LRT-NG and GoTube, evaluating their performance and identifying their characteristics and limits in verifying continuous-time systems with increasing complexity. We ran our experiments on a standard hardware setup (12 vCPUs, 64GB memory) equipped with a single GPU with a per-run timeout of 1 hour (except for runtimes reported in Figure \ref{fig:pareto}).
\begin{figure*}
     \centering
     \includegraphics[width=0.9\textwidth]{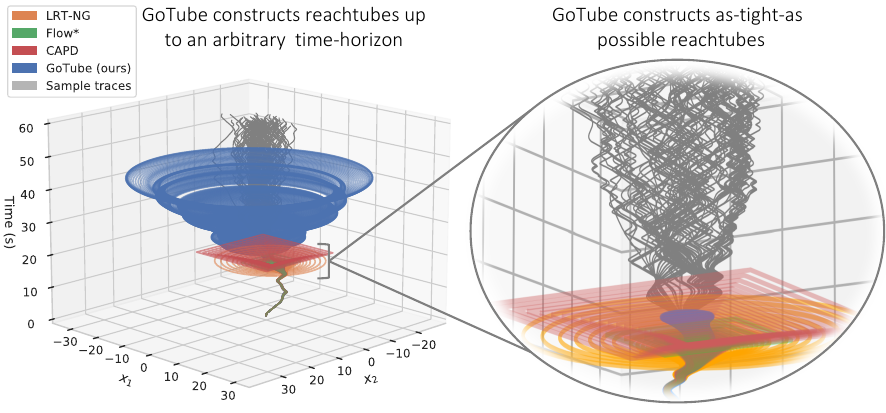}
        \caption{Visualization of the reachtubes constructed for the Dubin's car model with various reachability methods. While the tubes computed by existing methods (LRT-NG, Flow* and CAPD) explode at $t\,{\approx}\,20s$ (this moment is shown on the right side of the figure, where GoTube's reachtube is still very close to the sample traces) due to the accumulation of overapproximation errors (the infamous wrapping effect), GoTube can keep tight bounds beyond $t\,{>}\,40s$ for a 99\% confidence level (using 20000 samples, $\mu=1.1$ and runtime of one hour). Note also the chaotic nature of 100 executions. }
        \label{fig:three graphs}
\end{figure*}
\vspace{-5ex}
\subsection{On the Volume of the Bounding Balls}
Our first experimental evaluation concerns the overapproximation errors of the constructed bounding tubes.
An ideal reachability tool should be able to output an as tight as possible tube that encloses the system's executions. Consequently, as our comparison metric, we will report the average volume of the bounding balls, where less volume is better.
We use the benchmarks and settings of \cite{gruenbacher2020lagrangian} (same radii, time horizons, and models) as the basis of our evaluation. In particular, we compare GoTube to the deterministic, state-of-the-art reachability tools LRT-NG, Flow*, CAPD, and LRT. We measure the volume of GoTube's balls at the confidence levels of 90\% and 99\%, using $\mu=1.1$ as the tightness factor (in the third experiment we will talk in more detail about the trade-off between tightness and runtime).

\begin{figure}
    \centering
    \includegraphics[width=0.5\textwidth]{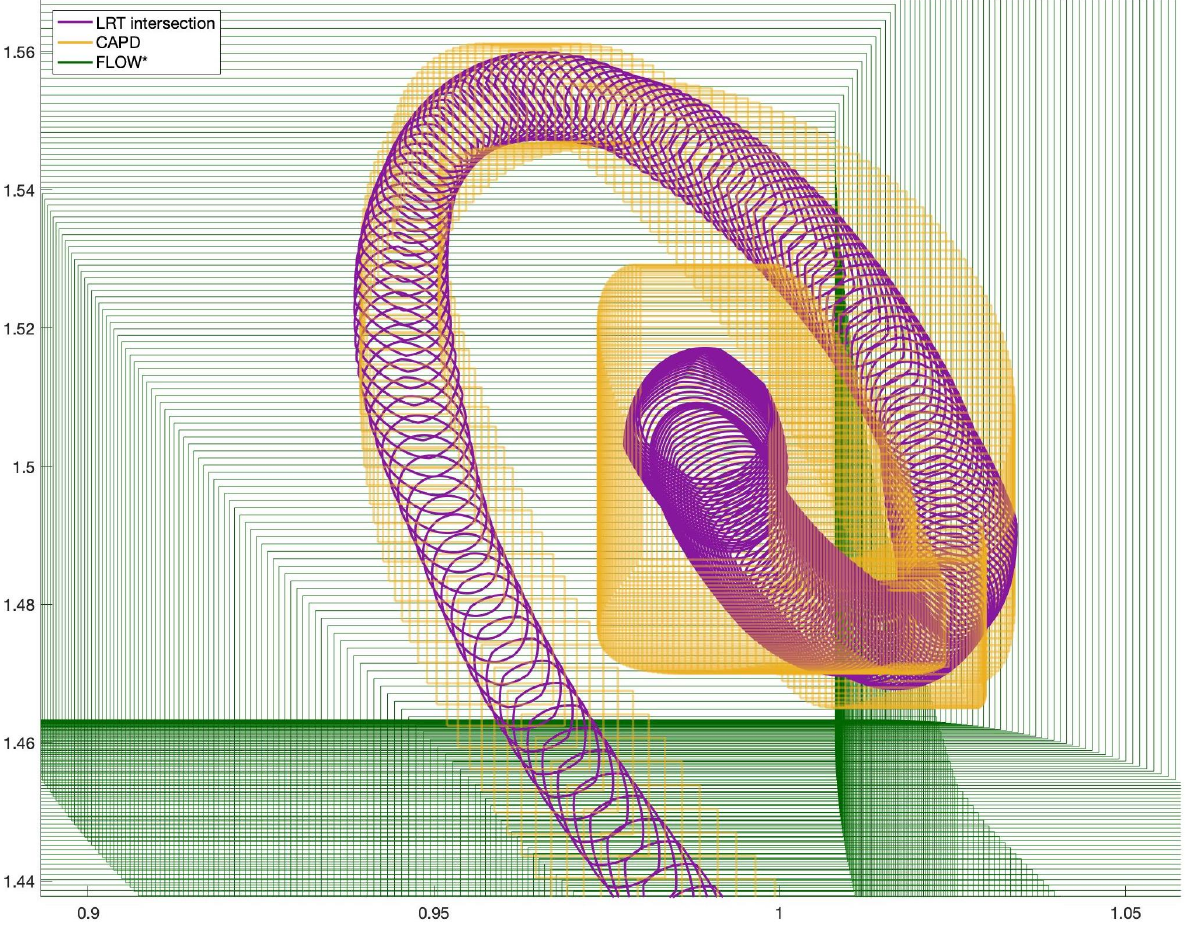}
    \caption{Comparison of the conservative reachtubes computed by the Flow* (in green), CAPD (in orange), and LRT-NG (in violet). Flow* blows up already before time $t=10$, while CAPD works up to $t=19.1$ and LRT-NG up to $t=22.6$. As clearly shown in this figure of the reachtube until time $t=14$, LRT-NG is superior to both Flow* and CAPD on this Brusselator model.}
    \label{fig:Brusselator}
    \vspace*{-3ex}
\end{figure}

The results are shown in Table \ref{tab:volume}, and exemplary in Fig.~\ref{fig:Brusselator}. For the first five benchmarks, which are classical dynamical systems, we use the small time horizons $T$ and small initial radii $\rd_0$, which the other tools could handle.

LRT-NG is the only conservative tool able to handle the continuous-depth models. GoTube, with 99\% confidence, achieves a competitive performance to the other tools, coming out on top in 3 out of 5 benchmarks - using $\mu=1.1$ as the tightness bound. Intuitively this means, we are confident that the overapproximation includes all executions with a confidence level $1-\lambda$, but this overapproximation might not be as tight as desired. 

GoTube is able to achieve any desired tightness by reducing $\mu$ and increasing the runtime. The specific reachtubes and the chaotic nature of hundred executions of Dubin's car are shown in Figure \ref{fig:three graphs}. As one can see, the GoTube reachtube extends to a much longer time horizon, which we fixed at 40s. All other tools blew up before 20s. For the two problems involving neural networks, GoTube produces significantly tighter reachtubes.  
\begin{table}[t]
    \centering
    \caption{Comparison of LRT-NG and GoTube (using tightness bound $\mu=1.1$) to existing reachability methods. The first five benchmarks concern classical dynamical systems, whereas the two bottom rows correspond to time-continuous RNN models (LTC $=$ liquid time-constant networks) in a closed feedback loop with an RL environment \cite{Hasani2021liquid,vorbach2021causal}. The numbers show the volume of the constructed tube. Lower is better; best number in given in boldface.    }
    \begin{tabular}{l|llll|ll}
    \toprule \multirow{2}{*}{Benchmark} & \multirow{2}{*}{LRT-NG} &\multirow{2}{*}{Flow*}  &\multirow{2}{*}{CAPD}  & \multirow{2}{*}{LRT} & \multicolumn{2}{c}{GoTube}\\
         &    &   &  &  & (90\%) & (99\%)\\
       \midrule
       Brusselator  & 1.5e-4 & 9.8e-5  & 3.6e-4 & 6.1e-4 & 8.6e-5 & \textbf{8.6e-5} \\
       Van Der Pol &  4.2e-4 & \textbf{3.5e-4} & 1.5e-3  &  3.5e-4 & 3.5e-4 &  \textbf{3.5e-4} \\
       Robotarm & \textbf{7.9e-11} & 8.7e-10  & 1.1e-9 & Fail &  2.5e-10 & 2.5e-10 \\
       Dubins Car & 0.131  & 4.5e-2 & 0.1181 & 385 &  2.5e-2 & \textbf{2.6e-2} \\
       Cardiac Cell &  \textbf{3.7e-9} & 1.5e-8 & 4.4e-8  & 3.2e-8 &   4.2e-8 &  4.3e-8 \\
       \midrule
       CartPole-v1+LTC  & 4.49e-33  & Fail & Fail & Fail & 2.6e-37 &  \textbf{4.9e-37}  \\
       CartPole-v1+CTRNN  & 3.9e-27  & Fail & Fail & Fail &  9.9e-34 &  \textbf{1.2e-33}   \\
      \bottomrule
    \end{tabular}
    \label{tab:volume}
\vspace{-3ex}
\end{table}
\vspace{-2ex}
\subsection{GoTube Provides Safety Bounds up an Arbitrary Time Horizon} 
In our second experiment, we evaluate for how long LRT-NG, GoTube and existing methods, can construct a reachtube before exploding due to overapproximation errors. To do so, we extend the benchmark setup by increasing the time horizon for which the tube should be constructed, use tightness bound $\mu=1.1$ and set a 95\% confidence level, that is, probability that $\max_{\{x_1,\dots,x_m\}\subset\calB_0} d_j(x)$ is for any $m$-dimensional set and any time $t_j$ smaller than the bounding ball's radius.

The results in Table~\ref{tab:time} demonstrate that LRT-NG is the only conservative tool able to run on continuous-depth models and that GoTube produces significantly longer reachtubes than all considered state-of-the-art approaches, without suffering from severe overapproximation errors. Fig.~\ref{fig:intro} visualizes the difference to the existing methods and gives over-approximation margins, for two example dimensions of the CartPole-v1 environment and its CT-RNN controller.
\begin{table}[ht]
\small
    \centering
    \caption{Results of the extended benchmark by longer time horizons. The numbers show the volume of the constructed tube, ``Blowup'' indicates that the method produced \texttt{Inf} or \texttt{NaN} values due to a blowup. Lower is better; the best method is shown in bold.}
    \begin{tabular}{l|cc|cc}\toprule
         Benchmark  & \multicolumn{2}{c|}{CartPole-v1+CTRNN} & \multicolumn{2}{c}{CartPole-v1+LTC} \\
         Time horizon  & 1s & 10s & 0.35s & 10s \\\midrule
         LRT & Blowup & Blowup & Blowup & Blowup \\
         CAPD & Blowup & Blowup & Blowup & Blowup \\
         Flow* & Blowup & Blowup & Blowup & Blowup\\
         LRT-NG & 3.9e-27 & Blowup & 4.5e-33 & Blowup\\
         GoTube (ours) & \textbf{8.8e-34} & \textbf{1.1e-19} & \textbf{4.9e-37} & \textbf{8.7e-21} \\\bottomrule
         \end{tabular}
    \label{tab:time}
    \vspace*{-4ex}
\end{table}
\begin{figure*}[t]
    \centering
    \includegraphics[width=0.8\textwidth]{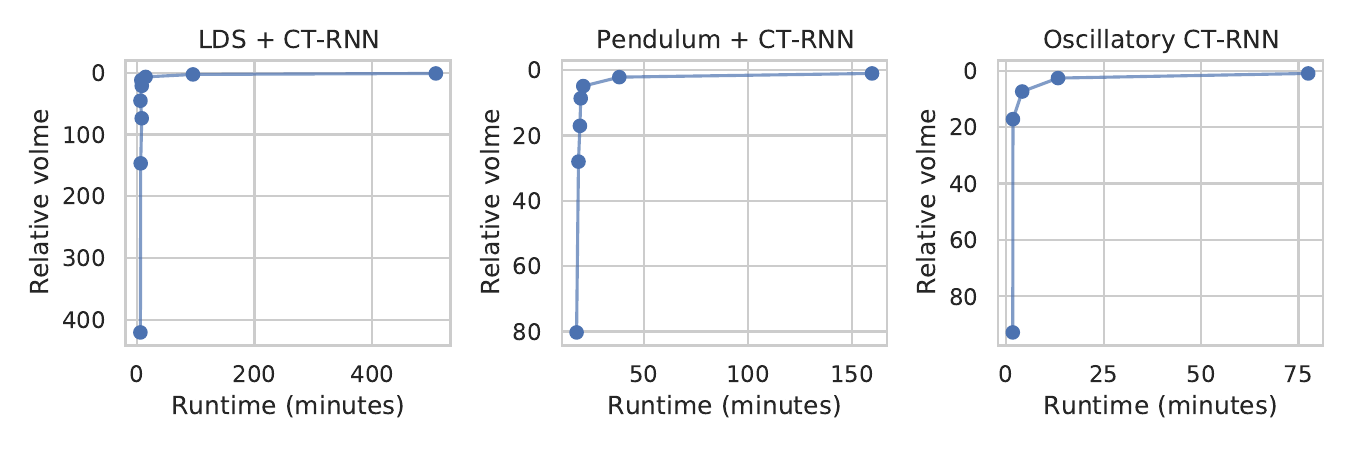}
    \caption{GoTube's runtime (x-axis) and volume size (y-axis) as a function of the tightness factor $\mu$. The volume was normalized by the volume obtained with the lowest $\mu$ (4.3e-13, 2.4e-12, and 2.1e-38 in particular).}
    \label{fig:pareto}
\end{figure*}
\subsection{GoTube Can Trade Runtime for Reachtube Tightness}
In our last experiment in~\cite{Gruenbacher2021GoTube}, we introduced a new set of benchmark models entirely based on continuous-time recurrent neural networks. 
The first model is an unstable linear dynamical system of the form $\dot{x} = Ax + Bu$ that is stabilized by a CT-RNN policy via actions $u$. The second model corresponds to the inverted pendulum environment, which is similar to the CartPole environment but differs in that the control actions are applied via a torque vector on the pendulum directly, instead of moving a cart. The CT-RNN policies for these two environments were trained using deep RL.
Our third new benchmark model concerns the analysis of the learned dynamics of a CT-RNN trained on supervised data. In particular, by using the reachability frameworks, we aim to assess if the learned network expressed oscillatory behavior. The CT-RNN state vector consists of 16 dimensions, which is twice as much as existing reachability benchmarks \cite{gruenbacher2020lagrangian}. 

Here, we study how GoTube can trade runtime for the volume of the constructed reachtube through its tightness factor $\mu$.
In particular, we run GoTube on our newly proposed benchmark with various values of $\mu$. We then plot GoTube's runtime (x-axis) and volume size (y-axis) as a function of $\mu$. The resulting curves show the Pareto-front of runtime-volume trade-off with GoTube.

Figure \ref{fig:pareto} shows the results for a time horizon of 10s in the first two examples, and of 2s in the last example.
Our results demonstrate that GoTube can adapt to different runtime and tightness constraints and set a new benchmark for future methods to compare with.
\section{Comparison of Both Approaches}\label{sec:comparison}
A common aspect of LRT, LRT-NG, SLR, and GoTube, is that they all make use of the variational Equation~\eqref{eq:variational equation}, together with the mean value Theorem~\ref{thm:mean value}. They allow our algorithms to have tighter bounds, less wrapping effect, and to be more efficient than other tools, as shown in the experimental evaluation. It is nevertheless important to know that for each tool, we had to develop new theoretical methods,  avoiding the blowing up either in space or in time.

When computing conservative guarantees, as in LRT-NG~\cite{gruenbacher2020lagrangian}, we employ the propagated interval deformation gradient, using the interval version of the variational Equations~\eqref{eq:interval variational equation}, and multiply the starting radius $\rd_0$ with the resulting stretching factor $\|\calF_j\|_{M_j}$ to over-approximate the set of reachable states at time $t_j$. As we need to use $[\calX_{j-1}]$ to compute $\|\calF_j\|_{M_j}$, the theoretical contributions of optimal metric computation and balls intersection with ellipsoids, are responsible for being the only conservative tool that can also verify continuous-depth models, by avoiding the accumulation of small errors (wrapping effect).

For the theoretical stochastic version of Lagrangian Reachability (SLR~\cite{Gruenbacher2021verification}), we use the variational equations even in two different ways: 1)~To propagate the deformation gradient for several samples - using the forward mode adjoint sensitivity method - to calculate the gradient of loss needed to find local minima. 2)~To propagate the interval variational equations and use Thm.~\ref{thm:mean value} to compute an upper bound of the Lipschitz constants for the distance function $d_j(x)$. This upper bound is used to compute the safety region radiuses.

In our scalable statistical robustness analysis tool GoTube, we completely avoid the use of interval arithmetic, as the interval Lipschitz constant in SLR leads to a blow-up in time. In Algorithm~\ref{algorithm:GoTube}, the variational equation is used to compute $F_x(t)$ via the forward mode adjoint sensitivity method for a tensorized batch of samples. Using Thm.~\ref{thm:mean value}, we compute local Lipschitz constants $\lambda_x$ for the samples, which we use to compute the cap radiuses and thus the probability.

Instead of using Lipschitz constants as a bloating factor for the ball's radius as in LRT-NG, we use it to define regions (caps) around already visited points on the surface, and to compute an upper bound for the values inside that caps - either deterministic safety regions (SLR) or statistical Lipschitz caps (GoTube). This knowledge allows us to compute the probability of having an upper bound for the global maximum of Eq.~\eqref{eq:optim}. The bigger the Lipschitz constant, the smaller the safety region radius and thus the probability. So a huge difference between the conservative and the statistical method is that a too large upper bound of the Lipschitz constant results in a state explosion for LRT-NG but in a time explosion for SLR and GoTube.

Another difference is that LRT-NG always computes as-tight-as-possible reachtubes, given the dynamical system. In contrast, SLR and GoTube allow to trade between time and accuracy, by using the tightness bound parameter $\mu$.
Thus, after finishing our global search strategy for timestep $t_j$, we have the statistical guarantee that the functional values of every $x\in\calB_0$ are less or equal to $\mu\cdot\bar{m}$. This implies that we should initiate the search with a relatively large $\mu=\mu_1$, obtaining for every $x$ a relatively large value of $r_{x,\mu_1}$ and therefore obtain a faster coverage of the search space. Subsequently, we can investigate whether the reachset $\calB_j$ with radius $\rd_j=\mu_1\cdot\bar{m}$ intersects with a region of bad (unsafe) states. If this is not the case, we can proceed to the next timestep $t_{j+1}$. Otherwise, we reduce $\mu$ to $\mu_2 < \mu_1$.
Accordingly, we can find a first radius for $\calB_j$ faster and refine it as long as $\calB_j$ intersects with the region of bad states.
\section{Conclusions and Future Work}\label{sec:conclusion}
In this work, we considered the robustness analysis of continuous-depth models to ensure safety of closed-loop cyber-physical systems with a neural network controller. We showed how to achieve tight reachtubes with deterministic (LRT-NG) or with statistical (SLR and GoTube) guarantees. We also compared the methods theoretically, by showing their common grounds in mathematical theory, and their distinct usage of that basis. As our experiments show, LRT-NG is superior to LRT, CAPD, and Flow*. Moreover, GoTube is stable and sets the state-of-the-art in terms of its ability to scale to time horizons well beyond what has been previously possible. Lastly, LRT-NG's and GoTube's scalability enables them to readily handle the verification of advanced continuous-depth neural models, a setting where state-of-the-art deterministic approaches fail.

Our current algorithms and methods require the availability of an ODE model of the environment. However, this is often not the case in complex applications, such as in autonomous driving, or in individual artificial pace makers. The ultimate goal is to scale up to high dimensional continuous-depth models and complex tasks, without knowing the model of the environment in advance.

%
%
%
\bibliographystyle{splncs04}
\bibliography{bibliography}
\end{document}